\documentclass{article}

\usepackage{arxiv}

\usepackage[utf8]{inputenc} % allow utf-8 input
\usepackage[T1]{fontenc}    % use 8-bit T1 fonts
\usepackage{hyperref}       % hyperlinks
\usepackage{url}            % simple URL typesetting
\usepackage{booktabs}       % professional-quality tables
\usepackage{amsfonts}       % blackboard math symbols
\usepackage{nicefrac}       % compact symbols for 1/2, etc.
\usepackage{microtype}      % microtypography
\usepackage{lipsum}
\usepackage{orcidlink, soul, hyperref}
\usepackage{color, xcolor, tcolorbox, amsmath, amssymb, amsfonts, subcaption, mwe, float, placeins} 
\usepackage{graphicx}
\graphicspath{ {./images/} }

\fancypagestyle{fancyfirst}{
  \fancyhead{}
  \fancyfoot{}
  \fancyfoot[C]{
    \parbox{\textwidth}{
      \centering
      \textbf{Accepted for publication in:} \textit{European Conference on Machine Learning and Principles and Practice of Knowledge Discovery in Databases (ECML PKDD 2025). Research Track}\\
      \textbf{Editors:} Rita P. Ribeiro, Bernhard Pfahringer, Pedro Larrañaga, and Nathalie Japkowicz\\
      \textbf{To appear in:}  Springer in the Lecture Notes in Computer Science Series (LNCS), September 2025\\
      \thepage
    }
  }
}

\title{Advancing Multi-Step Mathematical Reasoning in Large Language Models through Multi-Layered Self-Reflection with Auto-Prompting}

\author{
 Andr\'e de Souza Loureiro\orcidlink{0009-0009-6856-2804} \\
  Luiz de Queiroz College of Agriculture \\
  University of São Paulo\\
  Brazil \\
  \texttt{a.loureiro@usp.br} \\
   \And
 Jorge Valverde-Rebaza\thanks{Corresponding author: jvalverr@tec.mx}\orcidlink{0000-0001-8664-9692} \\
  School of Engineering and Sciences \\
  Tecnologico de Monterrey\\
  Mexico \\
  \texttt{jvalverr@tec.mx} \\
  \And
 Julieta Noguez\orcidlink{0000-0002-6000-3452}  \\
  School of Engineering and Sciences \\
  Tecnologico de Monterrey\\
  Mexico \\
  \texttt{jnoguez@tec.mx} \\
\And
   David Escarcega\orcidlink{0009-0003-7374-3526}  \\
  School of Engineering and Sciences \\
  Tecnologico de Monterrey\\
  Mexico \\
  \texttt{descarcega@tec.mx} \\
\And
Ricardo Marcacini\orcidlink{0000-0002-2309-3487}\\
Institute of Mathematics and Computer Sciences\\
University of São Paulo\\
Brazil \\
\texttt{ricardo.marcacini@usp.br}\\
}

\begin{document}
\maketitle
\thispagestyle{fancyfirst}

\begin{abstract}

Recent advancements in Large Language Models (LLMs) have significantly improved their problem-solving capabilities. However, these models still struggle when faced with complex multi-step reasoning tasks. In this paper, we propose the \emph{Multi-Layered Self-Reflection with Auto-Prompting} (MAPS) framework, a novel approach designed to enhance multi-step mathematical reasoning in LLMs by integrating techniques such as Chain of Thought (CoT), Self-Reflection, and Auto-Prompting. Unlike traditional static prompting methods, MAPS employs an iterative refinement process. Initially, the model generates a solution using CoT prompting. When errors are detected, an adaptive self-reflection mechanism identifies and analyzes them, generating tailored prompts to guide corrections. These dynamically adjusted prompts enable the model to iteratively refine its reasoning. Experiments on four well-established benchmarks across multiple LLMs show that MAPS significantly outperforms standard CoT and achieves competitive results with reasoning-optimized models. In addition, MAPS enables general-purpose LLMs to reach performance levels comparable to specialized reasoning models. While deeper reflection layers improve accuracy, they also increase token usage and costs. To balance this trade-off, MAPS strategically limits reflection depth, ensuring an optimal balance between cost and reasoning performance.

\end{abstract}

%%%%%%%%%%%%%%%%%%%%%%%%%%%%%%%%%%%%%%%%%%%%%%%%%%%%%%%%%%%%%%%%
%%%%%%%%%%%%%%%%%%%%%%%%%%%%%%%%%%%%%%%%%%%%%%%%%%%%%%%%%%%%%%%%

\section{Introduction}
\label{sec:intro}

Large Language Models (LLMs) have significantly impacted a wide range of applications, including healthcare, finance, education, and others~\cite{yu:nlp_reasoning_24,minaee:llm-survey:2024}. Despite these advances, researchers in academia and industry continue striving to equip LLMs with human-like reasoning skills to enhance generalization in real-world problem-solving through abstraction and logical inference~\cite{xu:reasoning_survey:2025,Valmeekam:LRMs:2024,yu:nlp_reasoning_24}.

 A common approach involves fine-tuning LLMs for logical and mathematical tasks. For instance, models from the GPT-4 family~\cite{openai:gpt4:2024} have demonstrated strong performance in logical inference, problem-solving, and mathematical reasoning. Models such as OpenAI o3-mini~\cite{openai:o3_mini:2025} and DeepSeek-R1~\cite{deepseek:2025}, further strengthen reasoning and coding. However, while effective in multi-step problem-solving, pre-trained models with native reasoning abilities require substantial computational resources, making training and deployment costly. 

A more resource-efficient alternative to extensive pre-training is prompt tuning, a process in which a pre-trained model is further optimized using curated datasets containing labeled instruction-response pairs~\cite{Wang:zero:CoT:2023}. However, conventional prompting strategies such as zero-shot or auto-prompts do not fully exploit the reasoning potential of LLMs~\cite{minaee:llm-survey:2024,xu:reasoning_survey:2025}.

Recent advances in adaptive prompting techniques have aimed to enhance multi-step reasoning. Chain-of-Thought (CoT) prompting, for example, guides the model to generate intermediate reasoning steps before arriving at a final answer~\cite{Wei:CoT:2022}. Although CoT improves performance, it does not always prevent the propagation of errors. Self-Reflection (SR) has been introduced to address this shortcoming by prompting the model to critically review and adjust its own responses, mimicking human self-correction~\cite{Renze:SelfReflection:2024}. However, relying solely on single-pass reflection often limits the model’s ability to correct deeper logical or arithmetic mistakes. Thus, for more complex problem statements, multiple iterative reflection layers are needed to achieve better results~\cite{Mirzadeh:GSM:2024}.

These limitations underscore the need for more sophisticated techniques capable of iteratively refining a model's reasoning process. To address this, we propose the \emph{\textbf{M}ulti-layer \textbf{A}uto-\textbf{P}rompted \textbf{S}elf-reflection} (MAPS) framework, a novel approach designed to enhance reasoning capabilities by dynamically generating customized reflection prompts and incorporating iterative feedback mechanisms. In contrast to conventional prompting techniques that utilize static reflection prompts, MAPS engages in a multi-stage process. Initially, the model generates a preliminary solution using CoT prompting, which guides the reasoning process through explicit step-by-step analysis. If the initial answer is found to be incorrect, the framework then initiates adaptive reflection iterations. In these iterations, the model produces tailored prompts that specifically address the identified errors, allowing for focused self-reflection and correction. This iterative refinement process enables the model to improve its reasoning over successive attempts, ultimately leading to more accurate solutions.

To validate the effectiveness of the MAPS framework, we perform a comprehensive evaluation on GSM8K~\cite{Cobbe:GSM8k:2021}, GSM-Symbolic~\cite{Mirzadeh:GSM:2024}, AIME 2025~\cite{balunovic:srimatharena_aime:2025}, and MATH~\cite{Hendrycks:math_dataset:2021} datasets. The results demonstrate that MAPS significantly enhances the ability of general-purpose LLMs, such as LLaMA, to detect and correct errors. Furthermore, MAPS demonstrates competitive performance when compared to pre-trained models that are specifically engineered with inherent reasoning capabilities, such as OpenAI's o3-mini and o4-mini~\cite{openai_o3_o4mini_systemcard:2025}, as well as Google's Gemini 2.5 Flash and Gemini 2.5 Pro~\cite{gemini:family_highly_capable:25}.

The remainder of this paper is structured as follows. Section~\ref{sec:related_work} provides a review of the most relevant prompting techniques for enhancing reasoning in large language models. Section~\ref{sec:proposal:MAPS} introduces the MAPS framework, detailing its operational mechanisms. Section~\ref{sec:experimental} outlines the experimental setup designed to assess the performance of MAPS. Section~\ref{sec:results} presents and discusses the results of the experimental evaluation across various models. Finally, Section~\ref{sec:conclusion} summarizes the key findings and highlights potential directions for future research.

%%%%%%%%%%%%%%%%%%%%%%%%%%%%%%%%%%%%%%%%%%%%%%%%%%%%%%%%%%%%%%%%
%%%%%%%%%%%%%%%%%%%%%%%%%%%%%%%%%%%%%%%%%%%%%%%%%%%%%%%%%%%%%%%%

\section{Related Work}
\label{sec:related_work}

General-purpose LLMs often struggle with solving mathematical word problems. This challenge arises in part because transformer-based architectures are inherently designed to generate text one token at a time~\cite{Cobbe:GSM8k:2021}. Consequently, to improve their problem-solving capabilities, it is crucial to design prompts that encourage step-by-step reasoning. 

A key approach to guiding LLMs in step-by-step reasoning is prompt learning, which instructs the model to follow structured reasoning steps. Wei et al. (2022)~\cite{Wei:CoT:2022} introduced Chain-of-Thought (CoT) prompting, showing that prompting an LLM to rephrase question information as intermediate steps significantly improves performance over direct answers. CoT’s success has driven further research into reasoning in LLMs, inspiring techniques like auto-CoT, ZS-CoT, Complexity-based prompting, Tree of Thoughts (ToT), and others~\cite{xu:reasoning_survey:2025,zhong:gsm8k:2024}.

Renze \& Guven (2024)~\cite{Renze:SelfReflection:2024} introduced Self-Reflection (SR) prompting to address CoT’s limitations. While CoT improves reasoning through step-by-step guidance, it lacks mechanisms for evaluating and correcting errors. SR prompting enables models to refine responses, enhancing accuracy and problem-solving. However, its effectiveness depends on the use of well-constructed prompts~\cite{liu:selfreflectionoutcomesensitiveprompt:2024} and the implementation of multi-layered strategies to tackle more complex logical or arithmetic problems~\cite{Mirzadeh:GSM:2024}.

Despite advancements, existing methods often struggle with dynamic error adaptation in complex tasks. Techniques like CoT and SR provide structured reasoning but lack iterative error correction. Our proposal addresses these gaps by integrating adaptive self-reflection to identify errors and generate tailored prompts, enhancing reasoning in general-purpose LLMs and bridging the gap between static methods and dynamic reasoning needs.

%%%%%%%%%%%%%%%%%%%%%%%%%%%%%%%%%%%%%%%%%%%%%%%%%%%%%%%%%%%%%%%%
%%%%%%%%%%%%%%%%%%%%%%%%%%%%%%%%%%%%%%%%%%%%%%%%%%%%%%%%%%%%%%%%

\section{MAPS: Multi-layer Auto-Prompted Self-reflection}
\label{sec:proposal:MAPS}

In this section, we present our novel framework, \emph{Multi-Layered Self-Reflection with Auto-Prompting} (MAPS), which is designed to enhance multi-step reasoning in LLMs. MAPS builds on the Chain-of-Thought (CoT) paradigm and traditional self-reflection techniques by introducing an iterative mechanism that dynamically adjusts reflection prompts according to the problem's structure, complexity, and previously identified errors. This adaptive approach seeks to improve the model's reasoning accuracy and robustness by fostering deeper introspection and targeted problem-solving strategies. The details of MAPS are explained below.

\subsection{Framework Overview}
\label{sec:proposal:MAPS:overview}

\begin{figure}[!ht]
\centering
\begin{subfigure}{0.48\textwidth} 
    \centering
    \scalebox{1.38}{
\includegraphics[trim=2cm 1.7cm 0cm 1.1cm, clip, width=0.85\columnwidth]{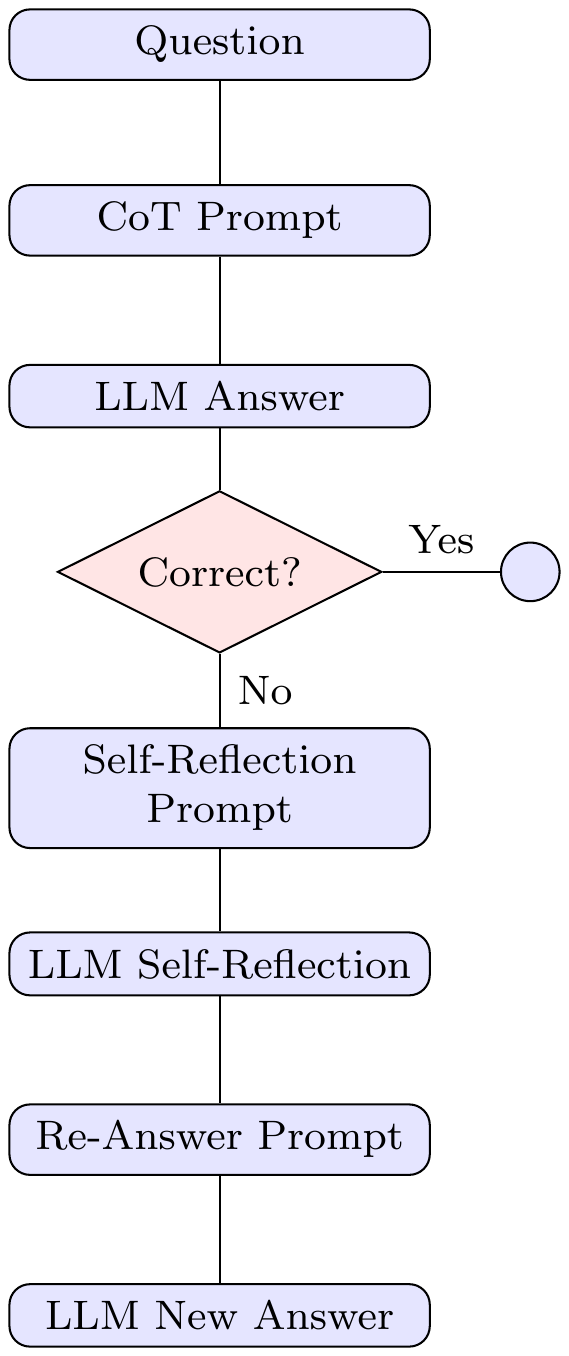}
    }
    \caption{SR with single-pass.} 
    \label{fig:singlepass}
\end{subfigure}
\hfill
\begin{subfigure}{0.48\textwidth}
    \centering
    \scalebox{1.38}{
\scalebox{0.98}{
\hspace{-0.4cm}\includegraphics[trim=1.7cm 0.9cm 0.2cm 0cm, clip, width=0.85\columnwidth]{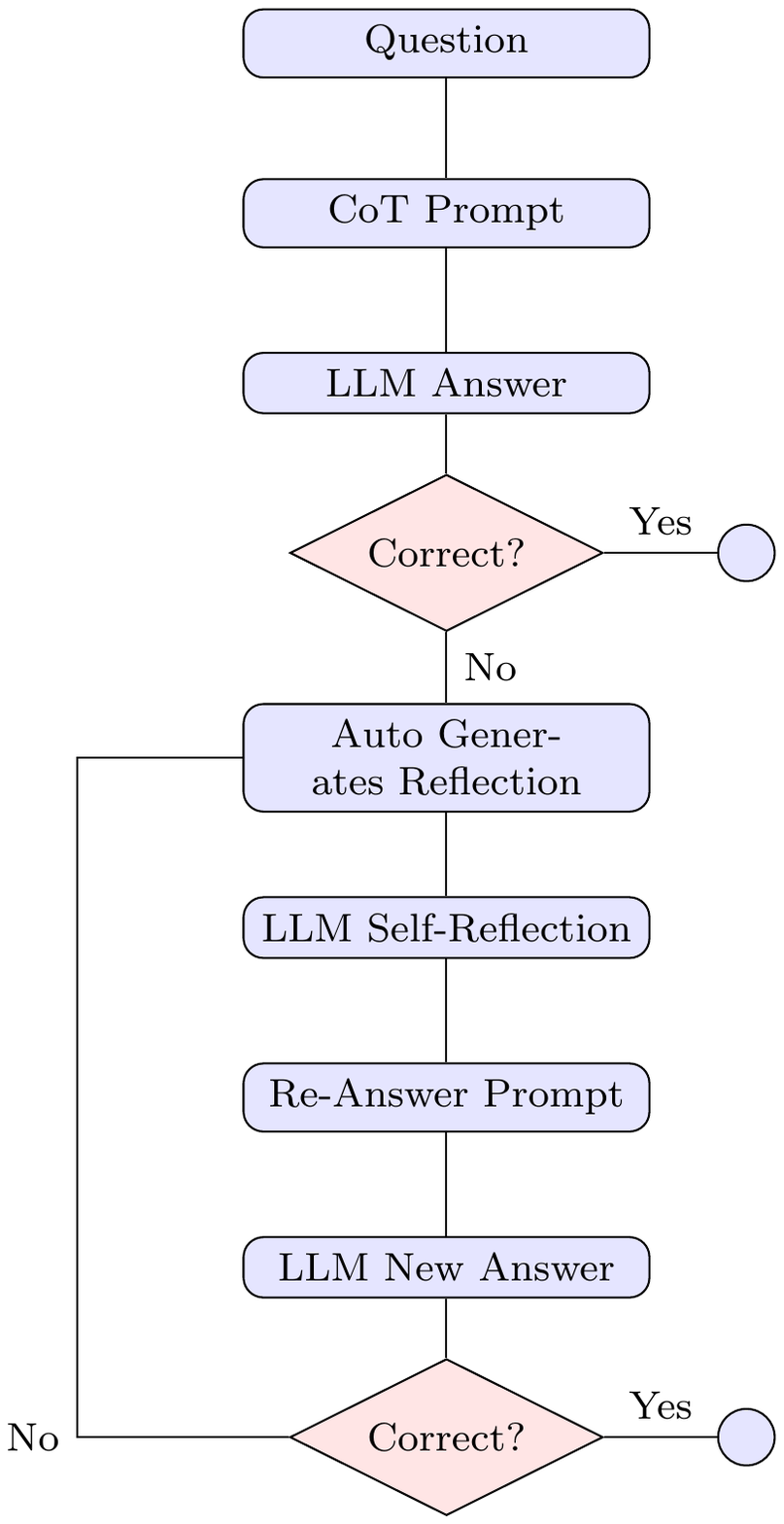}
    }
    }
    \caption{MAPS.} 
    \label{fig:maps}
\end{subfigure}
\caption{Comparison of (a) Self-Reflection with single-pass and (b) Multi-layer Auto-Prompted Self-Reflection (MAPS).}
\label{fig:reflection_diagrams}
\end{figure}

Traditional prompting methods, including the original version of SR (\textit{i.e.}, single-pass SR), utilize a fixed reflection prompt to identify and correct errors in an initial CoT-generated solution, as illustrated in Figure~\ref{fig:singlepass}. However, such static prompts may fail to address diverse error types or deeper logical and arithmetic mistakes, particularly in complex symbolic problems. To overcome these limitations, our framework introduces the following key contributions.

\begin{enumerate}
    \item Iterative Reflection: After the initial CoT response, the model’s answer is examined for correctness. If incorrect, the framework initiates one or more reflection iterations.
    
    \item Auto-Prompting (Meta-Prompting): Instead of applying a one-size-fits-all reflection template, the model is guided to generate a tailored self-reflection prompt. This reflection prompt is dynamically created based on the problem’s characteristics, known error patterns, and the complexity of the task.
    
    \item Dynamic Adaptation: If a single reflection iteration does not yield the correct answer, additional layers of self-reflection are executed. In each iteration, a new auto-generated reflection prompt is produced to iteratively refine the reasoning until the correct solution is obtained or a preset maximum number of iterations is reached.
\end{enumerate}

Therefore, while SR performs one reflection cycle using a static prompt, MAPS fine-tunes the model iteratively by generating customized reflection prompts and continuously updating answers, as depicted in Figure~\ref{fig:maps}. This capability allows for systematic detection and correction of errors, ensuring better performance in reasoning-related tasks.

\begin{figure}[ht]
    \centering
    \includegraphics[trim=0cm 5.6cm 0cm 0cm, clip, width=0.85\columnwidth]{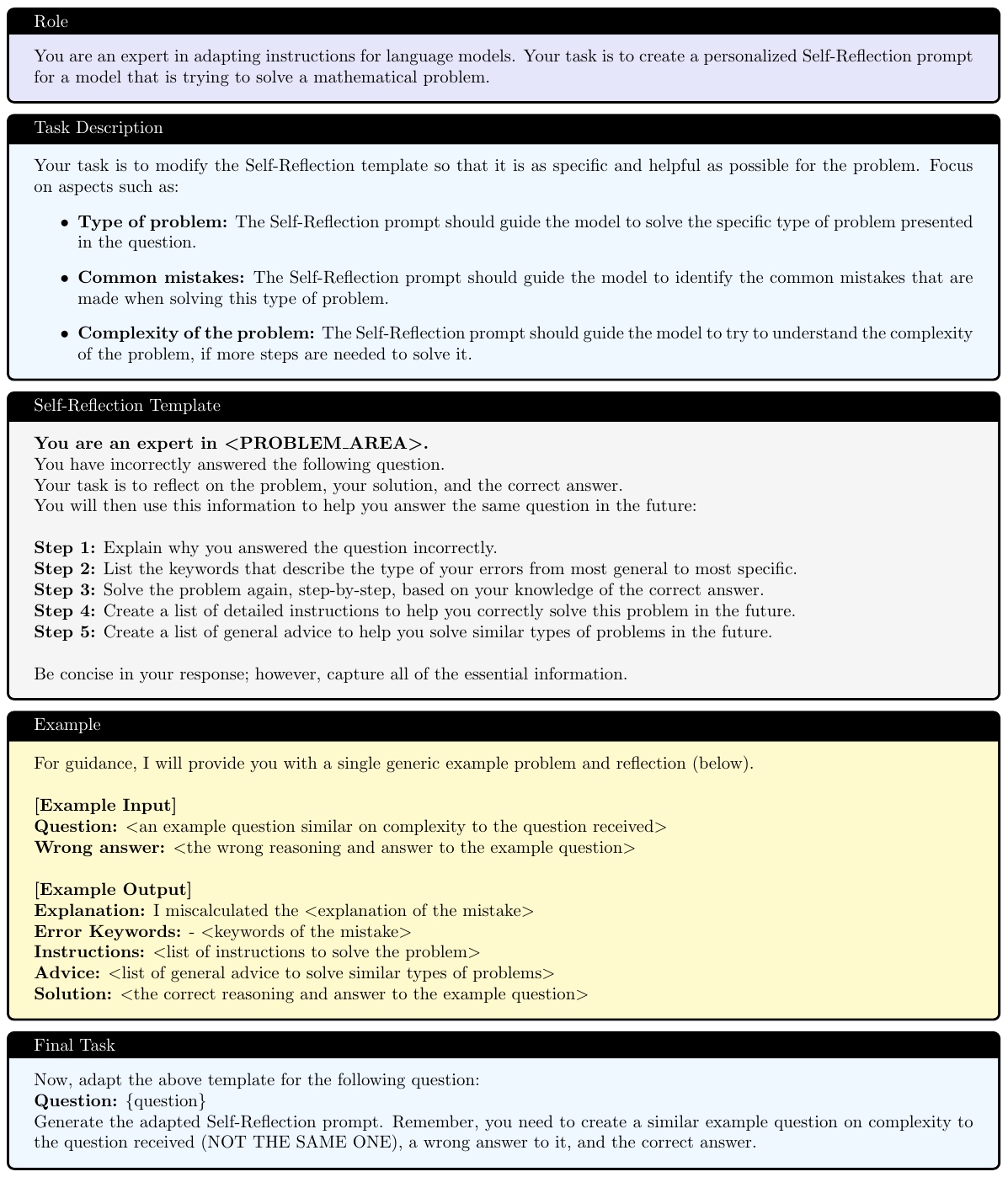}
    \caption{Meta-prompt (template) to apply MAPS.}
    \label{fig:maps-template}
\end{figure}

\subsection{Methodology}
\label{sec:proposal:MAPS:methodology}
As illustrated in Figure~\ref{fig:maps}, our approach proceeds in the following steps:

\begin{enumerate}
    \item Initial CoT Reasoning: The LLM is provided with the original question and a Chain-of-Thought prompt (\textit{e.g.}, “\textit{Let’s think step by step} \ldots”). This produces an initial answer along with intermediate reasoning steps.
    
    \item Correctness Verification: The output is evaluated against the expected answer or verified using an external correctness check. If the response is correct, the process is terminated.
    
    \item Auto-Prompt Generation: If the answer is incorrect, the LLM is tasked with generating a customized reflection prompt that adapts the standard reflection template to the specifics of the problem. This meta-prompt encourages the LLM to: (i) diagnose its mistakes, (ii) list common error types, and (iii) provide refined instructions for re-solving the problem. 
    
    \item Self-Reflection and Re-Answering: Guided by the auto-generated reflection prompt, the model analyzes the errors in its previous attempt, identifies the missteps, and then re-solves the problem with corrective feedback incorporated.
    
    \item Iterative Update: The newly generated answer undergoes verification. If it remains incorrect, the auto-prompt generation and self-reflection cycle repeat until a correct solution is produced or a predefined maximum number of iterations is reached. We recommend limiting this process to a maximum of three cycles (layers) to balance thoroughness with computational efficiency.
\end{enumerate}

By iteratively generating and responding to tailored auto-prompts, the model systematically identifies and corrects errors through self-reflection. Consequently, MAPS effectively stimulates the reasoning capabilities of base models.

\subsection{MAPS Meta-prompt}
\label{sec:proposal:MAPS:finetuning}

Figure~\ref{fig:maps-template} illustrates the meta-prompt that guides the model in generating tailored self-reflection prompts for each question. Rather than relying on static instructions, the MAPS meta-prompt defines the model's role as an expert in adapting reasoning strategies and ensures that self-reflection is dynamically adjusted to the specific characteristics of the problem. 

This adaptation within the meta-prompt allows the model to consider crucial factors such as the type of problem (e.g., arithmetic or geometry), typical errors associated with that domain, and the complexity of the reasoning required. The meta-prompt employs a structured yet flexible framework, enabling the model to integrate pertinent domain knowledge at each stage of reflection. To support this process, examples, including error cases, are provided to demonstrate effective self-reflection. Finally, the model applies this structured methodology to novel problems, generating reflection prompts dynamically rather than relying on static templates.

\subsection{Applying MAPS}
\label{sec:proposal:example}

\begin{figure}[H]
    \centering
    \includegraphics[trim=0cm 24.8cm 0cm 0cm, clip, width=0.85\columnwidth]{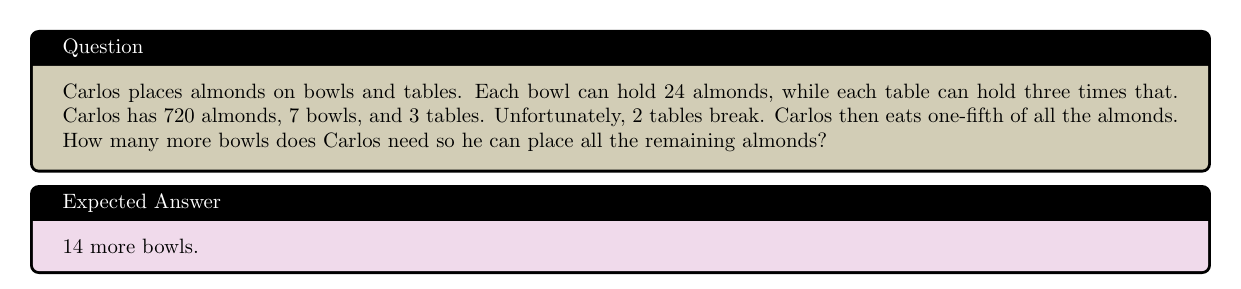}
    \caption{Instance of question and expected answer from \textit{GSM-symbolic-p2}.}
    \label{fig:qa-maps}
\end{figure}

MAPS can be applied to any LLM. To illustrate its use, we implement it on \texttt{Llama 3.1-8B Instruct} and demonstrate its performance on a representative instance from the \textit{GSM-symbolic-p2} dataset, which naturally embodies a mathematical problem (details in Section~\ref{sec:experimental:datasets}). The selected data instance is depicted in Figure~\ref{fig:qa-maps}.

The reflection cycle for this problem, executed through CoT and MAPS, is summarized in Table~\ref{tab:multi_layer_example}. Initially, the CoT reasoning produces an incorrect answer of $7$ more bowls, likely due to an error in adding or subtracting the available capacity. In the first reflection layer for MAPS ($1L$), despite the use of an auto-generated prompt, the LLM fails to identify and correct the miscalculation. However, in the second reflection layer ($2L$), an adapted prompt generated based on the prior error analysis enables the model to identify the mistake and compute the correct answer of $14$ more bowls.

\begin{table}[H]
\centering
\caption{Summary of multi-layer reflection iterations for a sample from \textit{GSM-symbolic-p2}.}
\label{tab:multi_layer_example}
\renewcommand{\arraystretch}{1.1}
\setlength{\tabcolsep}{4pt}
\begin{tabular}{lccc}
\hline
\textbf{Stage} & \textbf{Answer} & \textbf{Reflection Layer} & \textbf{Correct?} \\
\hline
\emph{CoT} & 7  & -- & No \\
\emph{MAPS 1L} & $7$  & $1$ & No \\
\emph{MAPS 2L} & $14$ & $2$ & Yes \\
\hline
\end{tabular}
\end{table}

\begin{figure}[ht]
    \centering
    \includegraphics[trim=0cm 19.9cm 0cm 0cm, clip, width=0.85\columnwidth]{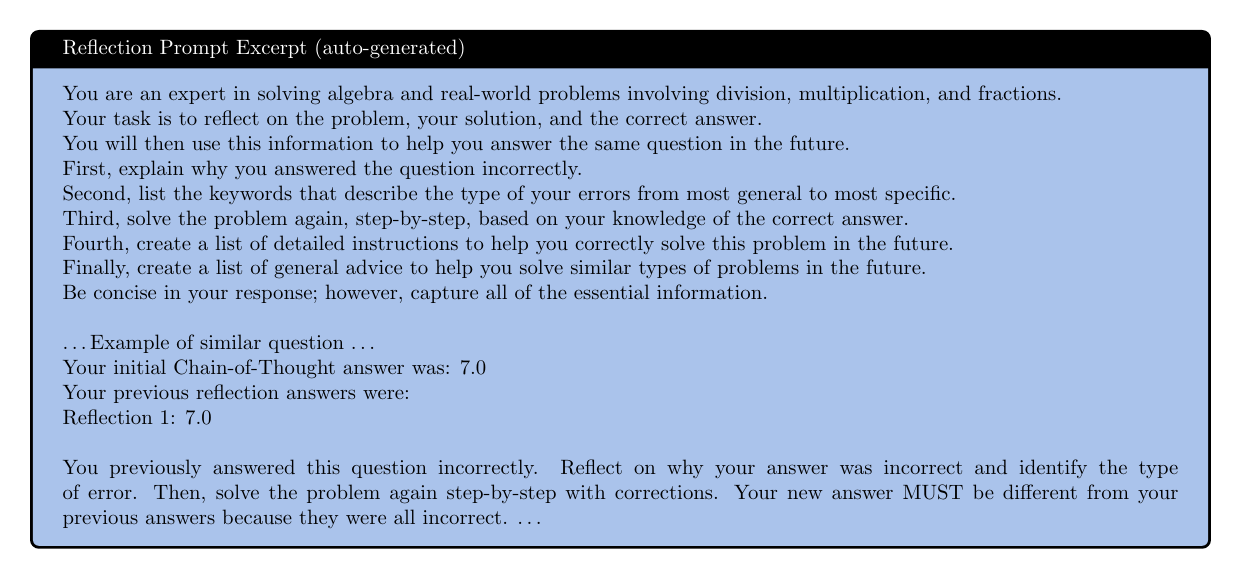}
    \caption{Excerpt of the auto-generated reflection prompt from the second MAPS cycle ($2L$).}
    \label{fig:reflection-prompt-maps}
\end{figure}

This example clearly illustrates the fundamental capability of MAPS: when a single-pass reflection is insufficient, multiple layers of auto-prompting and self-reflection progressively refine the solution until the correct answer is obtained. To provide a closer approximation of the model's reflection process, an excerpt from the second reflection prompt is presented in Figure~\ref{fig:reflection-prompt-maps}.

%%%%%%%%%%%%%%%%%%%%%%%%%%%%%%%%%%%%%%%%%%%%%%%%%%%%%%%%%%%%%%%%
%%%%%%%%%%%%%%%%%%%%%%%%%%%%%%%%%%%%%%%%%%%%%%%%%%%%%%%%%%%%%%%%

\section{Experimental Setup}
\label{sec:experimental}

This section outlines the experimental setup for evaluating MAPS in prompt tuning to enhance multi-step reasoning in LLMs, ensuring reproducibility and transparency.

\subsection{Models}
\label{sec:experimental:models}

Our evaluation encompassed a diverse array of  LLMs, selected based on variations in parameter size, architecture, computational efficiency, and reasoning capabilities.

Firstly, we evaluated smaller but cost-effective models such as Meta’s \texttt{Llama 3.1–8B Instruct}, OpenAI's \texttt{GPT-4o-mini-2024-07-18} and Google's \texttt{Gemma-2-} \texttt{9b-it}. These models are optimized for rapid inference and lower operational costs, although their performance on complex tasks tends to be limited. Second, we considered mid-sized models, including Google's \texttt{Gemma-2-27b-it} and Mistral AI’s instruct-optimized \texttt{Mistral-Small-24b-Instruct-2501}. These models strike an effective balance between computational expense and performance, demonstrating robust capabilities across a wide range of applications. Lastly, we examine larger models, such as Meta’s \texttt{Llama 3.1–70B Instruct}, DeepSeek’s \texttt{DeepSeek‑V3}, and OpenAI’s and \texttt{GPT‑4o‑2024‑11‑20}, which were selected for their superior performance enabled by extensive parameter counts and advanced pre‑training techniques.

To establish performance benchmarks, we included specialized reasoning models from OpenAI, such as \texttt{o1-preview}, \texttt{o3-mini}, and \texttt{o4-mini}, as well as Google's models, \texttt{Gemini 2.5 Flash} and \texttt{Gemini 2.5 Pro}. These models are explicitly optimized for complex reasoning tasks and serve as essential reference points for contextualizing the results obtained from general-purpose models enhanced by MAPS.

%%%%%%%%%%%%%%%%%%%%%%%%%%%%%%%%%%%%%%%%%%%%%%%%%%%%%%%%%%%%%%%
%%%%%%%%%%%%%%%%%%%%%%%%%%%%%%%%%%%%%%%%%%%%%%%%%%%%%%%%%%%%%%%
%%%%%%%%%%%%%%%%%%%%%%%%%%%%%%%%%%%%%%%%%%%%%%%%%%%%%%%%%%%%%%%

\subsection{Datasets}
\label{sec:experimental:datasets}

We evaluated our approach using four prominent datasets for mathematical reasoning: \textit{GSM8K}, \textit{GSM-Symbolic}, \textit{AIME 2025}, and \textit{MATH 500}, which collectively cover a broad spectrum of difficulty, from basic arithmetic to advanced symbolic manipulation, allowing for a comprehensive evaluation of our proposal.

\paragraph{\textbf{GSM8K.}} This dataset contains grade school-level mathematical problems requiring multi-step reasoning, serving as a standard benchmark for assessing LLM performance on basic arithmetic and algebra tasks~\cite{Cobbe:GSM8k:2021}.

\paragraph{\textbf{GSM-Symbolic.}} Derived from \textit{GSM8K}, this dataset introduces varying levels of symbolic complexity to assess reasoning robustness~\cite{Mirzadeh:GSM:2024}. It includes three progressively challenging variants. The \textit{main} variant consists of original problems with modified entity names and numerical values, designed to evaluate the model's ability to generalize beyond surface-level features. The \textit{p1} variant extends the \textit{main} version by adding an extra complexity clause, requiring more nuanced reasoning. Finally, the \textit{p2} variant represents the most challenging setting, incorporating two additional complexity clauses to test performance under significant symbolic transformations.

\paragraph{\textbf{AIME 2025.}} The American Invitational Mathematics Examination (AIME) consists of highly challenging competition-level problems in algebra, combinatorics, and number theory~\cite{balunovic:srimatharena_aime:2025}. These questions are crafted to test the boundaries of advanced mathematical reasoning, making AIME 2025 a formidable benchmark for evaluating current models.

\paragraph{\textbf{MATH 500.}} This dataset is a curated selection of 500 diverse problems from the original MATH benchmark, encompassing topics such as probability, algebra, trigonometry, and geometry~\cite{Hendrycks:math_dataset:2021}. It is designed to assess a model's capability to apply mathematical principles and execute complex calculations.

\subsection{Evaluation Metrics}
\label{sec:experimental:metrics}
To evaluate our framework, we used three key metrics:

\paragraph{\textbf{Accuracy.}} The proportion of correctly answered questions, \( N_{\text{corr}} \), relative to the total number of questions, \( N_{\text{total}} \). Thus, accuracy is given by \( \text{Accuracy} = \frac{N_{\text{corr}}}{N_{\text{total}}} \).

\paragraph{\textbf{Symbolic Loss.}} Measures the accuracy drop from the GSM8K dataset to the GSM-Symbolic variants, calculated as \( \text{Symbolic Loss} = \text{Accuracy}_{\text{GSM8K}} - \text{Accuracy}_{\text{GSM-Symbolic}} \). Lower values indicate greater robustness to symbolic complexity.

\paragraph{\textbf{Cost Analysis.}} The total inference cost per 100 questions, calculated from generated tokens and provider-specified costs.

%%%%%%%%%%%%%%%%%%%%%%%%%%%%%%%%%%%%%%%%%%%%%%%%%%%%%%%%%%%%%%%
%%%%%%%%%%%%%%%%%%%%%%%%%%%%%%%%%%%%%%%%%%%%%%%%%%%%%%%%%%%%%%%
%%%%%%%%%%%%%%%%%%%%%%%%%%%%%%%%%%%%%%%%%%%%%%%%%%%%%%%%%%%%%%%

\subsection{Procedure}
\label{sec:experimental:procedure}

This study conducts two evaluations to assess the effectiveness of prompting techniques for tuning both conventional LLMs and reasoning-specialized models. The first evaluation explores smaller, mid-sized, and larger general-purpose LLMs with limited reasoning abilities and apply the prompting methods:

\begin{enumerate}
    \item \textit{Baseline:} The model is provided solely with the problem statement, without additional instructions or step-by-step reasoning prompts (\textit{e.g.} zero-shot). For the MATH 500 and AIME 2025 datasets, however, the prompt includes a minimal instruction to format the final answer using boxed notation.
    
    \item \textit{Chain-of-Thought (CoT):} For the GSM8K and GSM-Symbolic datasets, this involves providing the model with eight exemplar problems that illustrate step-by-step reasoning, followed by the directive to “\textit{think step by step}”. For AIME 2025 and MATH 500 datasets, the CoT prompt instructs the model to reason step-by-step and present the final answer in boxed notation.
    
    \item \textit{Self-Reflection (SR) with Single-Pass:} Following the generation of an initial response based on CoT, a predetermined static reflection prompt is appended to the model's output.
    
    \item \textit{Multi-layered Adaptive Prompting Strategy (MAPS):} Our proposed framework, MAPS, begins with the initial CoT-derived answer and subjects it to a multi-round iterative self-reflection process. We consider scenarios where a single reasoning layer is employed (MAPS $1L$) as well as cases where the iterative process continues until either a correct answer is reached or a predefined limit of three reflection layers is attained (MAPS $2$-$3L$).
\end{enumerate}

For the second evaluation, we expanded our investigation by comparing the performance of MAPS on the previously evaluated LLMs with that of advanced reasoning models.

All experiments were conducted in Python using either OpenRouter's\footnote{\url{https://openrouter.ai/}} or OpenAI's API\footnote{\url{https://platform.openai.com/}}. To ensure output consistency and facilitate reproducibility, we fixed the temperature at $0$ and top\_p at $1$. Our evaluation protocol was adapted according to the scale of each benchmark. For the large-scale datasets GSM8K and GSM-Symbolic, we performed five independent runs, each using a distinct random sample of 100 questions, to ensure result robustness without compromising experimental feasibility. The final accuracy scores for these two benchmarks reflect the mean performance across the five samples. In contrast, for the smaller datasets AIME 2025 and MATH 500, evaluation was carried out on their entire test sets based on a single execution.

%%%%%%%%%%%%%%%%%%%%%%%%%%%%%%%%%%%%%%%%%%%%%%%%%%%%%%%%%%%%%%%%
%%%%%%%%%%%%%%%%%%%%%%%%%%%%%%%%%%%%%%%%%%%%%%%%%%%%%%%%%%%%%%%%

\section{Results and Discussion}
\label{sec:results}

In this section, we present the experimental results and have a comprehensive discussion of our findings.

\subsection{Accuracy Gains across Prompting Methods}
\label{sec:results:assessment1}

Table~\ref{tab:accuracy_table} reports the experimental accuracy for the prompting strategies defined in Section~\ref{sec:experimental:procedure}. We evaluated eight LLMs, categorized as smaller, mid-sized and larger, as mentioned in Section~\ref{sec:experimental:models}. For each dataset and LLM, the best result is highlighted in bold.

From Table~\ref{tab:accuracy_table}, we observe that across all models and datasets, step-by-step reasoning via CoT consistently enhances performance compared to the Baseline. The addition of SR further improves accuracy. Notably, the MAPS framework achieves the highest results, often surpassing SR even with just a single reflection layer (MAPS $1L$). When employing two to three reflection layers (MAPS $2-3L$), MAPS consistently delivers exceptional performance, particularly in the challenging GSM-Symbolic-p2 subset, where its iterative error-correction mechanism proves most effective. This trend is corroborated by results on the AIME 2025 and MATH 500 benchmarks, where MAPS $2-3L$ consistently attains the highest accuracy, highlighting its effectiveness in tackling complex mathematical reasoning tasks.

\begin{table}[h!]
\centering
\caption{Accuracy of different prompting strategies across all benchmarks. For GSM8K and GSM-Symbolic, results correspond to the mean accuracy over five independent runs, each based on a distinct random sample of 100 questions. Symbolic loss (in parentheses) indicates the performance drop relative to GSM8K. For AIME 2025 and MATH 500, results are based on a single evaluation over the full dataset.}

\label{tab:accuracy_table}
\renewcommand{\arraystretch}{1.1}
\setlength{\tabcolsep}{6pt}
\scalebox{0.68}{
\begin{tabular}{lccccc}
\hline
\textbf{Dataset} & \textbf{Base} & \textbf{CoT} & \textbf{SR} & \textbf{MAPS 1L} & \textbf{MAPS 2--3L} \\
\hline
% --- SMALL MODELS ---
\multicolumn{6}{c}{\textbf{\texttt{meta-llama/llama-3.1-8b-instruct}}} \\
GSM8K & 0.761 & 0.822 & 0.920 & 0.910 & \textbf{0.955} \\
GSM-Symbolic-main & 0.680 (-10.60\%) & 0.766 (-6.81\%) & 0.890 \textbf{(-3.26\%)} & 0.852 (-6.37\%) & \textbf{0.916} (-4.08\%) \\
GSM-Symbolic-p1 & 0.604 (-20.6\%) & 0.630 (-23.36\%) & 0.740 (-19.57\%) & 0.754 (-17.14\%) & \textbf{0.838} \textbf{(-12.25\%)} \\
GSM-Symbolic-p2 & 0.376 (-50.59\%) & 0.376 (-54.26\%) & 0.600 (-34.78\%) & 0.540 (-40.66\%) & \textbf{0.680} \textbf{(-28.80\%)} \\
AIME 2025 & 0.000 & 0.000 & 0.000 & 0.000 & 0.000 \\
MATH 500 & 0.470 & 0.360 & 0.410 & 0.470 & \textbf{0.540} \\
\hline
\multicolumn{6}{c}{\textbf{\texttt{google/gemma-2-9b-it}}} \\
GSM8K & 0.790 & 0.856 & 0.888 & 0.914 & \textbf{0.946} \\
GSM-Symbolic-main & 0.770 (-2.53\%) & 0.784 (-8.41\%) & 0.850 (-4.28\%) & 0.882 (-3.5\%) & \textbf{0.922} \textbf{(-2.54\%)} \\
GSM-Symbolic-p1 & 0.618 (-21.77\%) & 0.688 (-19.63\%) & 0.794 (-10.59\%) & 0.838 (-8.32\%) & \textbf{0.888} \textbf{(-6.13\%)} \\
GSM-Symbolic-p2 & 0.476 (-39.75\%) & 0.516 (-39.72\%) & 0.632 (-28.83\%) & 0.684 (-25.16\%) & \textbf{0.792} \textbf{(-16.28\%)} \\
AIME 2025 & 0.000 & 0.000 & 0.000 & 0.000 & 0.000 \\
MATH 500 & 0.440 & 0.420 & 0.430 & 0.500 & \textbf{0.520} \\
\hline
\hline
% --- MID-SIZED MODELS ---
\multicolumn{6}{c}{\textbf{\texttt{google/gemma-2-27b-it}}} \\
GSM8K & 0.822 & 0.950 & 0.976 & 0.972 & \textbf{0.986} \\
GSM-Symbolic-main & 0.778 (-5.35\%) & 0.846 (-10.95\%) & 0.878 (-10.04\%) & 0.910 (-6.38\%) & \textbf{0.940} \textbf{(-4.67\%)} \\
GSM-Symbolic-p1 & 0.756 (-8.03\%) & 0.900 (-5.26\%) & 0.938 (-3.89\%) & 0.942 (-3.09\%) & \textbf{0.956} \textbf{(-3.04\%)} \\
GSM-Symbolic-p2 & 0.660 (-19.71\%) & 0.784 (-17.47\%) & 0.860 (-11.89\%) & 0.872 (-10.29\%) & \textbf{0.936} \textbf{(-5.07\%)} \\
AIME 2025 & 0.033 & 0.000 & 0.000 & 0.000 & \textbf{0.067} \\
MATH 500 & 0.480 & 0.420 & 0.440 & 0.490 & \textbf{0.520} \\
\hline
\multicolumn{6}{c}{\textbf{\texttt{mistralai/mistral-small-3.1-24b-inst}}} \\
GSM8K & 0.858 & 0.970 & 0.972 & 0.980 & \textbf{0.980} \\
GSM-Symbolic-main & 0.804 (-6.29\%) & 0.928 (-4.33\%) & 0.942 (-3.09\%) & 0.948 (-3.27\%) & \textbf{0.962} \textbf{(-1.84\%)} \\
GSM-Symbolic-p1 & 0.748 (-12.82\%) & 0.898 (-7.42\%) & 0.920 (-5.35\%) & 0.952 (-2.86\%) & \textbf{0.974} \textbf{(-0.61\%)} \\
GSM-Symbolic-p2 & 0.716 (-16.55\%) & 0.768 (-20.82\%) & 0.840 (-13.58\%) & 0.884 (-9.8\%) & \textbf{0.948} \textbf{(-3.27\%)} \\
AIME 2025 & 0.033 & 0.033 & 0.033 & 0.067 & \textbf{0.100} \\
MATH 500 & 0.670 & 0.590 & 0.650 & 0.650 & \textbf{0.730} \\
\hline
\hline
% --- LARGE MODELS ---
\multicolumn{6}{c}{\textbf{\texttt{meta-llama/llama-3.1-70b-instruct}}} \\
GSM8K & 0.835 & 0.948 & 0.970 & 0.971 & \textbf{0.984} \\
GSM-Symbolic-main & 0.808 (-3.19\%) & 0.910 (-4.01\%) & 0.960 \textbf{(-1.03\%)} & 0.952 (-1.95\%) & \textbf{0.968} (-1.63\%) \\
GSM-Symbolic-p1 & 0.800 (-4.19\%) & 0.894 (-5.70\%) & 0.940 (-3.09\%) & 0.940 (-3.19\%) & \textbf{0.964} \textbf{(-2.03\%)} \\
GSM-Symbolic-p2 & 0.716 (-14.25\%) & 0.792 (-16.46\%) & 0.870 (-10.31\%) & 0.876 (-9.78\%) & \textbf{0.928} \textbf{(-5.69\%)} \\
AIME 2025 & 0.067 & 0.033 & 0.033 & 0.067 & \textbf{0.100} \\
MATH 500 & 0.580 & 0.560 & 0.600 & 0.590 & \textbf{0.660} \\
\hline
\multicolumn{6}{c}{\textbf{\texttt{deepseek/deepseek-V3}}} \\
GSM8K & 0.934 & 0.964 & 0.976 & 0.972 & \textbf{0.978} \\
GSM-Symbolic-main & 0.904 (-3.21\%) & 0.924 (-4.15\%) & 0.956 (-2.05\%) & 0.950 (-2.26\%) & \textbf{0.960} \textbf{(-1.84\%)} \\
GSM-Symbolic-p1 & 0.892 (-4.50\%) & 0.908 (-5.81\%) & 0.952 (-2.46\%) & 0.946 (-2.67\%) & \textbf{0.970} \textbf{(-0.82\%)} \\
GSM-Symbolic-p2 & 0.852 (-8.78\%) & 0.852 (-11.62\%) & 0.924 (-5.33\%) & 0.936 (-3.70\%) & \textbf{0.944} \textbf{(-3.48\%)} \\
AIME 2025 & 0.333 & 0.300 & 0.300 & \textbf{0.400} & \textbf{0.400} \\
MATH 500 & 0.750 & 0.760 & 0.760 & 0.780 & \textbf{0.810} \\
\hline
\multicolumn{6}{c}{\textbf{\texttt{gpt-4o-mini-2024-07-18}}} \\
GSM8K & 0.849 & 0.949 & 0.970 & 0.967 & \textbf{0.975} \\
GSM-Symbolic-main & 0.864 \textbf{(+1.77\%)} & 0.920 (-3.06\%) & 0.930 (-4.12\%) & 0.938 (-3.00\%) & \textbf{0.954} (-2.15\%) \\
GSM-Symbolic-p1 & 0.794 (-6.48\%) & 0.878 (-7.48\%) & 0.910 (-6.19\%) & 0.938 (-3.00\%) & \textbf{0.950} \textbf{(-2.56\%)} \\
GSM-Symbolic-p2 & 0.776 \textbf{(-8.60\%)} & 0.708 (-25.40\%) & 0.840 (-13.40\%) & 0.844 (-12.72\%) & \textbf{0.876} (-10.15\%) \\
AIME 2025 & \textbf{0.133} & 0.100 & 0.100 & 0.100 & \textbf{0.133} \\
MATH 500 & 0.660 & 0.630 & 0.640 & 0.670 & \textbf{0.700} \\
\hline
\multicolumn{6}{c}{\textbf{\texttt{gpt-4o-2024-11-20}}} \\
GSM8K & 0.844 & 0.943 & 0.956 & 0.969 & \textbf{0.979} \\
GSM-Symbolic-main & 0.810 (-4.03\%) & 0.908 (-3.71\%) & 0.924 (-3.35\%) & 0.928 (-4.23\%) & \textbf{0.952} \textbf{(-2.72\%)} \\
GSM-Symbolic-p1 & 0.782 (-7.35\%) & 0.944 \textbf{(+0.10\%)} & 0.964 (+0.84\%) & 0.970 \textbf{(+0.10\%)} & \textbf{0.980} \textbf{(+0.10\%)} \\
GSM-Symbolic-p2 & 0.830 (-1.66\%) & 0.915 (-3.00\%) & 0.956 (+0\%) & 0.976 \textbf{(+0.72\%)} & \textbf{0.984} (+0.51\%) \\
AIME 2025 & 0.067 & 0.100 & 0.100 & 0.133 & \textbf{0.167} \\
MATH 500 & 0.590 & 0.630 & 0.660 & 0.700 & \textbf{0.760} \\
\hline
\end{tabular}
}
\end{table}

\begin{figure}[h!]
    \centering
    \begin{subfigure}{0.39\textwidth}
        \includegraphics[trim=1.62cm 0cm 1cm 0cm, clip, width=\linewidth]{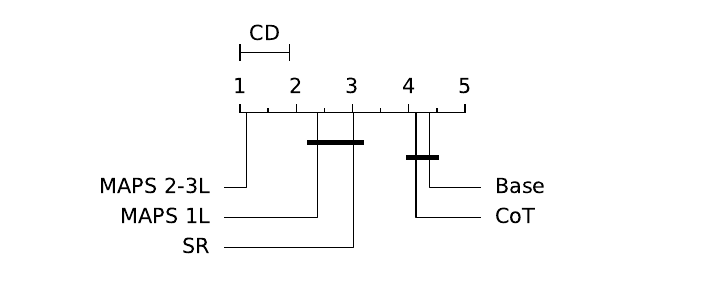}
        \caption{Prompting performance.}
        \label{fig:nemenyi_acc_prompting}
    \end{subfigure}
    \hfill
    \begin{subfigure}{0.59\textwidth}
        \includegraphics[trim=0.3cm 0cm 0.1cm 0cm, clip, width=\linewidth]{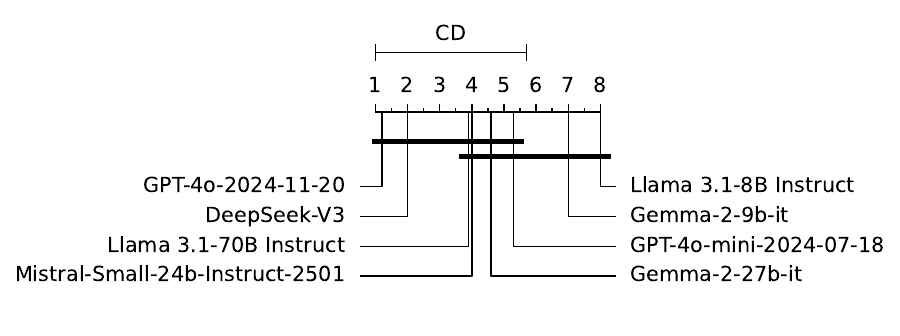}
        \caption{LLM performance.}
        \label{fig:nemenyi_acc_llms}
    \end{subfigure}
    \hfill
    \caption{Nemenyi post-hoc test applied to the accuracy results from Table~\ref{tab:accuracy_table} to statistically assess (a) the performance of prompting techniques and (b) their impact on LLM performance.}
    \label{fig:nemenyi_acc}
\end{figure}

To enhance the rigor of our analysis, we applied the Nemenyi post-hoc test~\cite{demsar:posthoctests:2006} to the accuracy results reported in Table~\ref{tab:accuracy_table}, with the corresponding critical difference diagram shown in Figure~\ref{fig:nemenyi_acc}. The Friedman test produced a statistic of $227.44$ with a p-value of $1.71 \times 10^{-25}$, indicating significant differences among the prompting strategies. The Nemenyi test identified a critical difference (CD) of $0.88$, confirming that the full version of MAPS, \textit{i.e.}, MAPS $2-3L$, is statistically superior to all other prompting methods. The basic version, MAPS $1L$, ranked second and showed no statistically significant difference from SR, which ranked third. Although CoT outperformed the Baseline, their performances remain statistically indistinguishable. Figure~\ref{fig:nemenyi_acc_prompting} visually summarizes the relative rankings among all evaluated strategies.

To better understand which of the evaluated LLMs benefit most from prompting strategies, we conducted the Nemenyi post-hoc test using only the accuracy results for the GSM-Symbolic-p2 subset presented in Table~\ref{tab:accuracy_table}. The Friedman test yielded a statistic of $30.48$ and a p-value of $3.91 \times 10^{-6}$, indicating statistically significant differences among the models. The subsequent Nemenyi test revealed a critical difference (CD) of $4.70$, as illustrated in Figure~\ref{fig:nemenyi_acc_llms}. The results show that larger models derive the greatest benefit from prompting techniques in terms of enhanced reasoning capabilities. Specifically, \texttt{GPT‑4o‑2024‑11‑20} and \texttt{DeepSeek‑V3} achieved the highest ranks, followed closely by \texttt{Llama 3.1–70B Instruct} in third and \texttt{GPT-4o-mini-2024-07-18} in sixth, with no statistically significant differences among them. Mid-sized LLMs such as \texttt{Mistral-Small-24b} \texttt{-Instruct-2501} and \texttt{Gemma-2-27b-it}, ranked fourth and fifth respectively, also demonstrated strong performance, statistically comparable to their larger counterparts. In contrast, the smaller models \texttt{Gemma-2-9b-it}, ranked seventh, and \texttt{Llama 3.1–8B Instruct}, ranked last, did not exhibit meaningful improvements in reasoning performance under the tested prompting methods.

%%%%%%%%%%%%%%%%%%%%%%%%%%%%%%%%%%%%%%%%%%%%%%%%%%%%%%
%%%%%%%%%%%%%%%%%%%%%%%%%%%%%%%%%%%%%%%%%%%%%%%%%%%%%%
%%%%%%%%%%%%%%%%%%%%%%%%%%%%%%%%%%%%%%%%%%%%%%%%%%%%%%

\subsection{Symbolic Loss Performance}
\label{sec:results:assessment1:symbolic_loss}

Still within the scope of the first evaluation procedure, Table~\ref{tab:accuracy_table} presents the symbolic loss values, shown in parentheses, calculated across the three \textit{GSM-Symbolic} variants and \textit{GSM8K}. The best symbolic loss values are highlighted in bold. Once again, MAPS $2-3L$ achieves the best performance in the vast majority of cases.

\vspace{-0.45cm}

The importance of symbolic loss lies in the fact that the \textit{GSM-Symbolic} benchmark introduces increasing symbolic complexity, progressing from \textit{main} to \textit{p1} and finally to \textit{p2}. As shown in Table~\ref{tab:accuracy_table}, each additional complexity clause significantly reduces accuracy for the Baseline, CoT, and SR techniques. In contrast, MAPS recovers much of the lost performance, demonstrating the value of iterative error diagnosis and correction for symbolic tasks.

Notably, certain larger models, such as \texttt{GPT-4o-2024-11-20}, can even surpass their \textit{GSM8K} accuracy on some symbolic variants when multi-layer reflection is applied, effectively achieving zero or even negative symbolic loss. For smaller and mid-sized models, such as \texttt{Llama 3.1--8B Instruct} and \texttt{Gemma-2-27b-it}, MAPS also mitigates symbolic loss effectively, though the gains are less pronounced compared to larger models, highlighting the influence of model capacity on the effectiveness of iterative prompting strategies.

%%%%%%%%%%%%%%%%%%%%%%%%%%%%%%%%%%%%%%%%%%%%%%%%%%%%%%
%%%%%%%%%%%%%%%%%%%%%%%%%%%%%%%%%%%%%%%%%%%%%%%%%%%%%%
%%%%%%%%%%%%%%%%%%%%%%%%%%%%%%%%%%%%%%%%%%%%%%%%%%%%%%

\subsection{MAPS versus specialized Reasoning Models}
\label{sec:results:assessment2}

In the context of the second evaluation procedure, as described in Section~\ref{sec:experimental:procedure}, we compare MAPS against specialized pre-trained LLMs renowned for their strong reasoning capabilities. Table~\ref{tab:specialized_reasoning_a} presents the comparison of general-purpose models enhanced with MAPS against native reasoning models such as \texttt{o1-Preview} and \texttt{o3-mini}, with the best accuracy results highlighted in bold.

\begin{table}[!ht]
\centering
\caption{Comparison of MAPS with specialized reasoning models: general-purpose LLMs + MAPS vs. native reasoning LLMs.}
\label{tab:specialized_reasoning_a}
\renewcommand{\arraystretch}{1.1}
\setlength{\tabcolsep}{4.5pt}
\scalebox{0.78}{
\begin{tabular}{lccccc}
\hline
\textbf{Benchmark} & \textbf{\shortstack{GPT-4o-2024-11-20 \\ + MAPS}} & 
\textbf{\shortstack{DeepSeek-V3 \\ + MAPS}} & 
\textbf{\shortstack{Llama 3.1--70B \\ + MAPS}} & 
\textbf{o1-preview} & \textbf{o3-mini} \\
\hline
GSM8K & \textbf{0.979} & 0.978 & 0.984 & 0.960 & 0.944 \\
GSM-Symbolic-main & 0.952 & 0.960 & \textbf{0.968} & 0.927 & 0.966 \\
GSM-Symbolic-p1 & \textbf{0.980} & 0.970 & 0.964 & 0.954 & 0.972 \\
GSM-Symbolic-p2 & \textbf{0.984} & 0.944 & 0.928 & 0.940 & 0.924 \\
\hline
\end{tabular}
}
\end{table}

\begin{figure}[ht]
    \centering
    \includegraphics[width=0.7\textwidth]{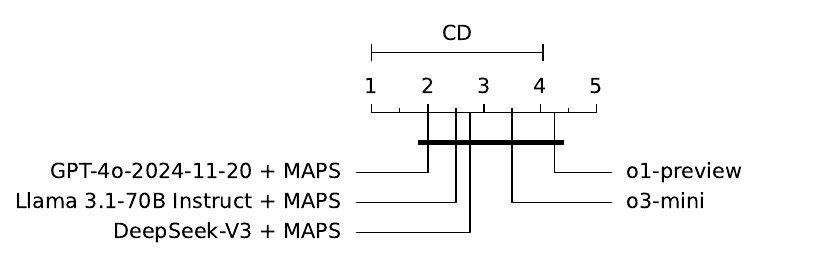}
    \caption{Nemenyi post-hoc test applied to the accuracy results of the top-3 general-purpose LLMs enhanced with MAPS against specialized reasoning models.}
    \label{fig:acc_reasoning}
\end{figure}

Overall, the MAPS approach enables several general-purpose models to match or even surpass specialized reasoning models, as evidenced in the upper section of Table~\ref{tab:specialized_reasoning_a}. For example, \texttt{GPT-4o-2024-11-20} with MAPS achieves 98.4\% accuracy on the challenging \textit{GSM-Symbolic-p2} subset, outperforming both \texttt{o1-Preview} and \texttt{o3-mini}. To assess the statistical significance of these differences, we applied the Friedman test, which yielded a statistic of $4.44$ and a p-value of $0.218$, indicating no significant performance differences among the compared models. The subsequent Nemenyi post-hoc test, with a critical difference (CD) of $3.05$, supports this conclusion. As illustrated in Figure~\ref{fig:acc_reasoning}, although the differences are not statistically significant, general-purpose LLMs augmented with MAPS consistently exhibit higher accuracy than their specialized counterparts under the same evaluation conditions.

To assess the robustness of our proposal, we applied MAPS to specialized reasoning models on the challenging AIME 2025 and MATH 500 benchmarks. As shown in Table~\ref{tab:specialized_reasoning_b}, MAPS consistently enhances performance across all models, including the state-of-the-art Gemini family. For example, on AIME 2025, \texttt{Gemini 2.5 Flash} improves from 66.7\% to 80.0\%, and \texttt{Gemini 2.5 Pro} from 80.0\% to 86.7\%. A similar +4.0\% absolute gain is observed on MATH 500 for both models. Additionally, OpenAI's \texttt{o4-mini} shows performance gains of up to +6.0\% on MATH 500.

\begin{table}[!ht]
\centering
\caption{Comparison of MAPS with specialized reasoning models: Applying MAPS to Reasoning LLMs on Complex Benchmarks.}
\label{tab:specialized_reasoning_b}
\renewcommand{\arraystretch}{1.1}
\setlength{\tabcolsep}{4.5pt}
\scalebox{0.78}{
\begin{tabular}{lccc}
\hline
\textbf{Benchmark} & \textbf{Model} & \textbf{CoT Accuracy} & \textbf{MAPS 2--3L} \\
\hline
AIME 2025 & \texttt{Gemini 2.5 Flash (no tools)} & 0.667 & \textbf{0.800} \\
AIME 2025 & \texttt{Gemini 2.5 Pro (no tools)}  & 0.800 & \textbf{0.867} \\
AIME 2025 & \texttt{o4-mini (medium / no tools)} & 0.800 & \textbf{0.867} \\
\hline
MATH 500 & \texttt{Gemini 2.5 Flash (05-20)} & 0.840 & \textbf{0.880} \\
MATH 500 & \texttt{Gemini 2.5 Pro (05-06) (no tools)}  & 0.840 & \textbf{0.880} \\
MATH 500 & \texttt{o4-mini (medium / no tools)} & 0.760 & \textbf{0.820} \\
\hline
\end{tabular}
}
\end{table}

These findings demonstrate that MAPS is a versatile enhancement strategy that improves inference quality across various models, even in competitive environments. This indicates that the core mechanism of MAPS offers a distinct advantage that complements the internal optimizations of modern LLMs.

%%%%%%%%%%%%%%%%%%%%%%%%%%%%%%%%%%%%%%%%%%%%%%%%%%%%%%
%%%%%%%%%%%%%%%%%%%%%%%%%%%%%%%%%%%%%%%%%%%%%%%%%%%%%%
%%%%%%%%%%%%%%%%%%%%%%%%%%%%%%%%%%%%%%%%%%%%%%%%%%%%%%

\subsection{Cost Analysis}

Table~\ref{tab:cost_comparison} presents the inference cost of MAPS on GSM-Symbolic for four representative general-purpose LLMs: \texttt{GPT-4o-mini-2024-07-18},  \texttt{GPT‑4o‑2024‑11‑20}, \texttt{Llama 3.1–8B Instruct}, and \texttt{Llama 3.1–70B Instruct}, in comparison to the specialized reasoning model \texttt{o3-mini}. While deeper reflection processes increase token usage and costs, so the results highlight key trade-offs in performance and efficiency.

The cost analysis in Table~\ref{tab:cost_comparison} shows that \texttt{GPT-4o-mini-2024-07-18} proves to be highly cost-effective, requiring only US\$ $0.045$ on \textit{main} and US\$ $0.08$ on \textit{p2}, costs significantly lower than \texttt{o3-mini}’s US\$ $0.078$ and US\$ $0.144$, respectively. Despite its lower cost, \texttt{GPT-4o-mini-2024-07-18} closely matches or slightly lags behind \texttt{o3-mini} in accuracy across most subsets. Although MAPS increases token usage due to auto-prompt generation and iterative refinement, \texttt{GPT-4o-mini-2024-07-18}’s low per-token cost offsets this, reinforcing its efficiency. \texttt{Llama-3.1-70B-Instruct} also delivers strong performance, matching or surpassing \texttt{o3-mini} in accuracy on certain tasks while remaining more economical, costing US\$ $0.052$ versus US\$ $0.078$ on \textit{main} (a $33$\% reduction) and US\$ $0.085$ versus US\$ $0.144$ on \textit{p2} ($41$\% lower). Like \texttt{GPT-4o-mini-2024-07-18}, we observe that \texttt{Llama-3.1-70B-Instruct} also benefits from lower inference costs, compensating for the added token consumption of deeper self-reflection layers. Conversely, \texttt{GPT-4o-2024-11-20} incurs the highest costs, reaching US\$ $0.94$ on \textit{main}, US\$ $0.92$ on \textit{p1}, and US\$  $1.07$ on \textit{p2}, over $600$\% higher than \texttt{o3-mini} despite outperforming it by $6$\% in accuracy on \textit{p2} ($0.984$ versus $0.924$).

%%%\vspace{-0.5cm}

\begin{table}[!ht]
\centering
\caption{Total inference cost (USD) for processing the \emph{GSM-Symbolic} dataset (100 questions).}
\label{tab:cost_comparison}
\renewcommand{\arraystretch}{1.1}
\setlength{\tabcolsep}{4pt}
\begin{tabular}{lccc}
\hline
\textbf{Model} & \textbf{Dataset} & \textbf{Total Cost (US\$)} \\
\hline
\textbf{GPT‑4o‑2024‑11‑20 + MAPS} & main & 0.944251 \\
                                        & p1   & 0.918467 \\
                                        & p2   & 1.066514 \\[3pt]
\textbf{GPT-4o-mini-2024-07-18 + MAPS} & main & 0.045059 \\
                                             & p1   & 0.053559 \\
                                             & p2   & 0.079530 \\[3pt]
\textbf{Llama 3.1–70B Instruct + MAPS}   & main & 0.052105 \\
                                             & p1   & 0.054528 \\
                                             & p2   & 0.084594 \\[3pt]
\textbf{Llama 3.1–8B Instruct + MAPS}    & main & 0.025040 \\
                                             & p1   & 0.034904 \\
                                             & p2   & 0.059944 \\[3pt]
\textbf{o3-mini}                              & main & 0.07795458 \\
                                             & p1   & 0.10527242 \\
                                             & p2   & 0.14444980 \\
\hline
\end{tabular}
\end{table}

%%%\vspace{-0.3cm}

The use of MAPS requires generating unique auto-prompts at each reflection layer, which increases token consumption and makes it less optimal for cost-sensitive applications. \texttt{Llama-3.1-8B-Instruct} is the most affordable, costing only US\$ $0.025$ on \textit{main}, but its accuracy in symbolic reasoning tasks is the lowest, ranking the worst in terms of performance. However, with a cost reduction of over $60$\% compared to \texttt{o3-mini} on \texttt{p2} (US\$ $0.060$ versus US\$ $0.144$), it presents a viable option for scenarios where minor accuracy trade-offs are acceptable. 

Overall, \texttt{GPT-4o-mini-2024-07-18} and \texttt{Llama-3.1-70B-Instruct} stand out as cost-efficient alternatives to reasoning-specific models like \texttt{o3-mini}, maintaining competitive performance at a fraction of the cost. While \texttt{GPT-4o-mini-2024-} \texttt{07-18} has considerable accuracy, its high cost makes it less attractive for practical deployment. Additionally, costs consistently increase from \textit{main} to \textit{p1} to \textit{p2} across all models, reflecting the additional complexity and token requirements associated with deeper reasoning steps in specialized models and extended self-reflection in our multi-layer approach.

%%%%%%%%%%%%%%%%%%%%%%%%%%%%%%%%%%%%%%%%%%%%%%%%%%%%%%
%%%%%%%%%%%%%%%%%%%%%%%%%%%%%%%%%%%%%%%%%%%%%%%%%%%%%%
%%%%%%%%%%%%%%%%%%%%%%%%%%%%%%%%%%%%%%%%%%%%%%%%%%%%%%

\subsection{Single-Pass Comparison: Traditional SR versus MAPS $1L$}
\label{sec:results:single_pass_comparison}

Before examining the benefits of multiple reflection layers, it is useful to compare the traditional SR method which considers single-pass with our proposal MAPS $1L$, \textit{i.e.} MAPS considering only one adaptive reflection layer. As observed in the SR and MAPS $1L$ results from Table~\ref{tab:accuracy_table}, it is possible to point out two key observations.

First, the overall performance of most LLMs remains similar under SR and MAPS $1L$ across \textit{GSM8K} and all variants of \textit{GSM-Symbolic}, suggesting that in a single-pass scenario, generating a custom reflection prompt in MAPS $1L$ does not significantly differ from using a well-tuned static prompt in SR. Second, model capacity plays a crucial role in the effectiveness of adaptive prompting. While \texttt{Llama-3.1-8B-Instruct} exhibits slightly lower accuracy with MAPS $1L$ compared to SR, particularly on \textit{p2}, indicating that smaller models may struggle to generate or leverage effective adaptive prompts in a single pass, \texttt{Mistral-Small-24b-Instruct-2501} consistently benefits from auto-prompted reflection, suggesting that certain architectures or parameter scales are better suited for adaptive meta-prompting. 

These findings suggest that auto-prompting in a single reflection pass can yield performance comparable to, or even surpass, that of a fixed self-reflection prompt. However, as evidenced by the results, incorporating additional reflection layers, as done in MAPS $2-3L$, consistently improves performance across all LLMs, highlighting the critical role of iterative error correction in addressing the most complex reasoning tasks.

%%%%%%%%%%%%%%%%%%%%%%%%%%%%%%%%%%%%%%%%%%%%%%%%%%%%%%%%%%%%%%%%
%%%%%%%%%%%%%%%%%%%%%%%%%%%%%%%%%%%%%%%%%%%%%%%%%%%%%%%%%%%%%%%%

\section{Conclusion}
\label{sec:conclusion}

This work introduced \emph{Multi-Layered Self-Reflection with Auto-Prompting} (MAPS), a framework that dynamically adapts reflection templates based on problem type and complexity. Through comprehensive evaluations on benchmarks such as GSM8K, GSM-Symbolic, AIME 2025, and MATH 500, MAPS consistently outperformed baseline methods including Chain-of-Thought (CoT) and single-pass Self-Reflection (SR). The most significant gains were observed on symbolic subsets \textit{p1} and \textit{p2}, underscoring the framework’s effectiveness in handling complex, abstract reasoning tasks.

MAPS enhances the reasoning capabilities of LLMs through an iterative and adaptive self-reflection process. This aligns with human-like problem-solving behaviors that rely on successive refinement and contextual learning. Notably, MAPS not only boosts general-purpose models to match or surpass specialized reasoning LLMs (such as \texttt{o1-Preview} and \texttt{o3-mini}), but also improves the performance of the specialized models themselves. For instance, state-of-the-art Gemini variants and OpenAI’s \texttt{o4-mini} exhibited consistent performance improvements when enhanced with MAPS, achieving average gains between 4\% and 6\% on challenging benchmarks like AIME 2025 and MATH 500. These findings confirm MAPS as a model-agnostic reasoning enhancer that generalizes well across architectures and difficulty levels.

Nonetheless, MAPS depends on the availability of explicit correctness signals to guide the self-reflection loop, which may restrict its applicability to open-ended tasks lacking ground-truth answers. Future work should address this limitation by integrating techniques such as uncertainty estimation, human-in-the-loop validation, or proxy supervision via auxiliary tasks.

Further research should also investigate the use of MAPS in other domains requiring structured reasoning, such as code generation, logical deduction, and scientific modeling. Enhancing the auto-prompting component with more expressive and context-sensitive mechanisms may further improve the system’s adaptability and efficiency. Overall, MAPS represents a promising direction toward more robust, generalizable, and cognitively inspired frameworks for LLM-based reasoning. 

To foster transparency and encourage further research, we release the full implementation of the MAPS framework, along with evaluation scripts and prompt templates, at: \url{https://github.com/and270/maps_prompting}.

%%%%%%%%%%%%%%%%%%%%%%%%%%%%%%%%%%%%%%%%%%%%%%%%%%%%%%%%%%%%%%%%
%%%%%%%%%%%%%%%%%%%%%%%%%%%%%%%%%%%%%%%%%%%%%%%%%%%%%%%%%%%%%%%%

\section*{Acknowledgement} 

The authors express their gratitude to the \textit{Fondo de Apoyo a Publicaciones} (FAP) of Tecnologico de Monterrey for the financial support. We also acknowledge the Department of Computing at the Tecnologico de Monterrey, Mexico City Campus, for providing the necessary resources and infrastructure.

%%%%%%%%%%%%%%%%%%%%%%%%%%%%%%%%%%%%%%%%%%%%%%%%%%%%%%%%%%%%%%%%
%%%%%%%%%%%%%%%%%%%%%%%%%%%%%%%%%%%%%%%%%%%%%%%%%%%%%%%%%%%%%%%%

\end{document}